\documentclass[letterpaper, 10pt, conference]{ieeeconf}  

\IEEEoverridecommandlockouts                              

\usepackage{multirow}
\usepackage{graphicx}
\usepackage{comment} 
\usepackage{balance}

\title{\LARGE \bf
Modeling Human Response to Robot Errors for Timely Error Detection
}

\author{Maia Stiber$^{1}$, Russell Taylor$^{1,2}$, and Chien-Ming Huang$^{1}$
\thanks{$^{1}$Dept. of Computer Science,
        Johns Hopkins University, Baltimore, Maryland, USA
        {\tt\small \{mstiber,rht,chienming.huang\}@jhu.edu}}%
\thanks{$^{2}$ Life Fellow, IEEE}%
}

\begin{document}

\maketitle
\thispagestyle{empty}
\pagestyle{empty}

\begin{abstract}
In human-robot collaboration, robot errors are inevitable---damaging user trust, willingness to work together, and task performance. Prior work has shown that people naturally respond to robot errors socially and that in social interactions it is possible to use human responses to detect errors. However, there is little exploration in the domain of non-social, physical human-robot collaboration such as assembly and tool retrieval. In this work, we investigate how people's organic, social responses to robot errors may be used to enable timely automatic detection of errors in physical human-robot interactions. 
We conducted a data collection study to obtain facial responses to train a real-time detection algorithm and a case study to explore the generalizability of our method with different task settings and errors. 
Our results show that natural social responses are effective signals for timely detection and localization of robot errors even in non-social contexts and that our method is robust across a variety of task contexts, robot errors, and user responses. This work contributes to robust error detection without detailed task specifications.


\end{abstract}

\section{Introduction}
Unmanaged robot errors are harmful to human-robot collaboration. These errors present a safety concern, damage task performance, and erode users' trust and willingness to continue that cooperative partnership~\cite{brooks2016analysis,salem2015would}. The first step towards successful error management is timely error detection, which is key to enabling error mitigation and recovery~\cite{yasuda2013psychological}. While prior research has explored various methods for error detection, these methods tend to be task dependent and are not adaptable to other contexts or unexpected errors~\cite{honig2021expect}. However, robot errors can be unexpected and independent of task, situation, and user~\cite{gideoni2022personal}. 

To enable automatic detection of robot errors that may occur in varying contexts, we build on prior work~\cite{stiber2020not} and explore how social signals may be an indicative source for automatic error detection. Though prior work has shown that human collaborators are likely to exhibit social signals due to errors' unexpectedness~\cite{giuliani2015systematic}, most research has been situated in social contexts using a social robot (e.g., \cite{kontogiorgos2021systematic,trung2017head, mirnig2017to}). It is, however, unclear whether this social signal-based approach to robot error detection would work for a non-social robot (e.g., a manipulator) interacting with people in a non-social setting (e.g., task demonstration).

\begin{figure}
  \includegraphics[width = \columnwidth]{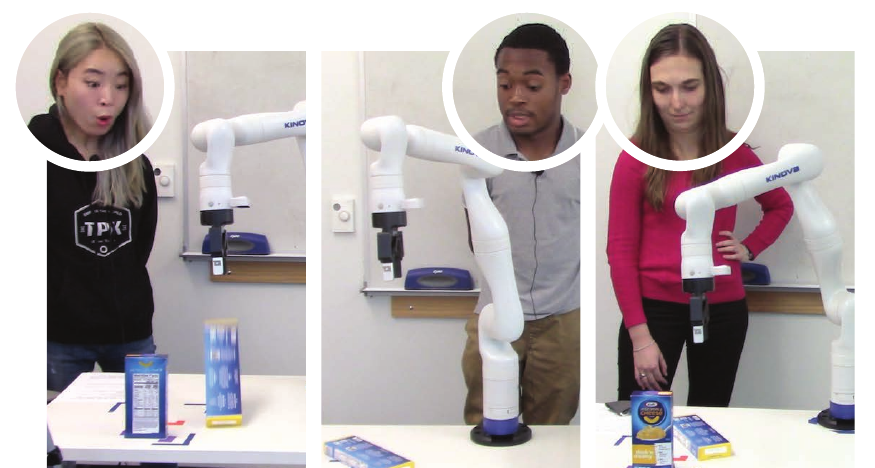}
\caption{In this work, we investigate the use of human response (action units) to detect unexpected robot errors in non-social interaction scenarios. 
Our approach works over a diverse array of tasks, errors, and responses.}
\label{fig:teaser}
\end{figure}

In this work, we explore how Action Units (AUs)---individual muscular movements as defined in the Facial Action Coding System~\cite{ekman1978facial}---may be used for automatic, timely detection of robot errors in non-social contexts. To this end, we first conducted a data collection study in which robot errors were intentionally produced to elicit natural social responses from participants (Fig. \ref{fig:teaser}). The collected data was then used to train a machine learning model for detection and temporal localization of robot errors. We additionally conducted a case study with a different set of participants experiencing different types of robot errors in various contexts to explore the generalizability of our approach. Our results indicate that AUs can be advantageously used for timely detection and temporal localization of unexpected robot errors and that our data-driven model can reasonably generalize its effectiveness to different settings and error types.
Our work makes the following set of contributions:
\begin{itemize}
    \item We show that it is possible to detect robot errors using social signals, especially AUs, in non-social, physical interaction scenarios.
    \item We develop a real-time human-in-the-loop error detection system, using the human collaborator as part of the detection process. 
    \item Our approach shows its effectiveness beyond its training setting and generalizes to different contexts and errors.
\end{itemize}

\section{Background and Related Work}
Robotic systems often make unavoidable technical and unexpected errors. For example, Nourbakhsh \textit{et al.} deployed three mobile robots in museums over five years and found that the mean time between failures was about 72 to 216 hours; it was difficult to increase the time between errors beyond that~\cite{nourbakhsh2003mobot}. Robot errors harm robot performance, which impacts user trust~\cite{schaefer2016meta} and the intensity of that negative effect is dependent on the error severity and quantity~\cite{rossi2017how,muir1996trust}. Furthermore, robot errors are dependent on the individual's perception and how a robot's behavior deviates from the individual's mental model of the robot~\cite{rossi2017human}.

\subsection{Social Responses to Robot Errors}
People respond socially to robot errors \cite{trung2017head,cahya2019static}. Moreover, humans exhibit more behaviors during error-occurring situations than error-free ones~\cite{cahya2019static}. In particular, gaze~\cite{kontogiorgos2021systematic,aronson2018gaze,kontogiorgos2020behavioral}, facial expressions~\cite{trung2017head}, verbalizations~\cite{kontogiorgos2021systematic}, and body movements~\cite{giuliani2015systematic,kontogiorgos2021systematic,bovo2020detecting} have been shown to be common instinctive responses to robot errors. Much of this prior work was contextualized in social scenarios and relied on participants’ existing expectations of robots during the interactions. Furthermore, the robots used in prior works typically were humanoid, which could have impacted the elicited responses~\cite{kontogiorgos2020behavioral}. Our preliminary work indicates that people exhibit social signals in response to robot errors even during physical human-robot interaction scenarios~\cite{stiber2020not}.

\subsection{Error Detection Using Social Signals}
To the best of our knowledge, there has been one system built utilizing social signals to detect social errors automatically. Kontogiorgos \textit{et al.} showed that---with a collection of head, body, gaze, AU, and verbal tracking---it is possible to both automatically detect and classify certain types of social errors during conversational failures with a social robot~\cite{kontogiorgos2021systematic}.
However, little is known about how social signal responses to robot errors may enable automatic error detection in physical human-robot interactions, such as a robot manipulator committing technical task errors. 

\section{Modeling User Response to Robot Errors}

\begin{figure*} \vspace{2.5mm}
\centering
  \includegraphics[width = \textwidth]{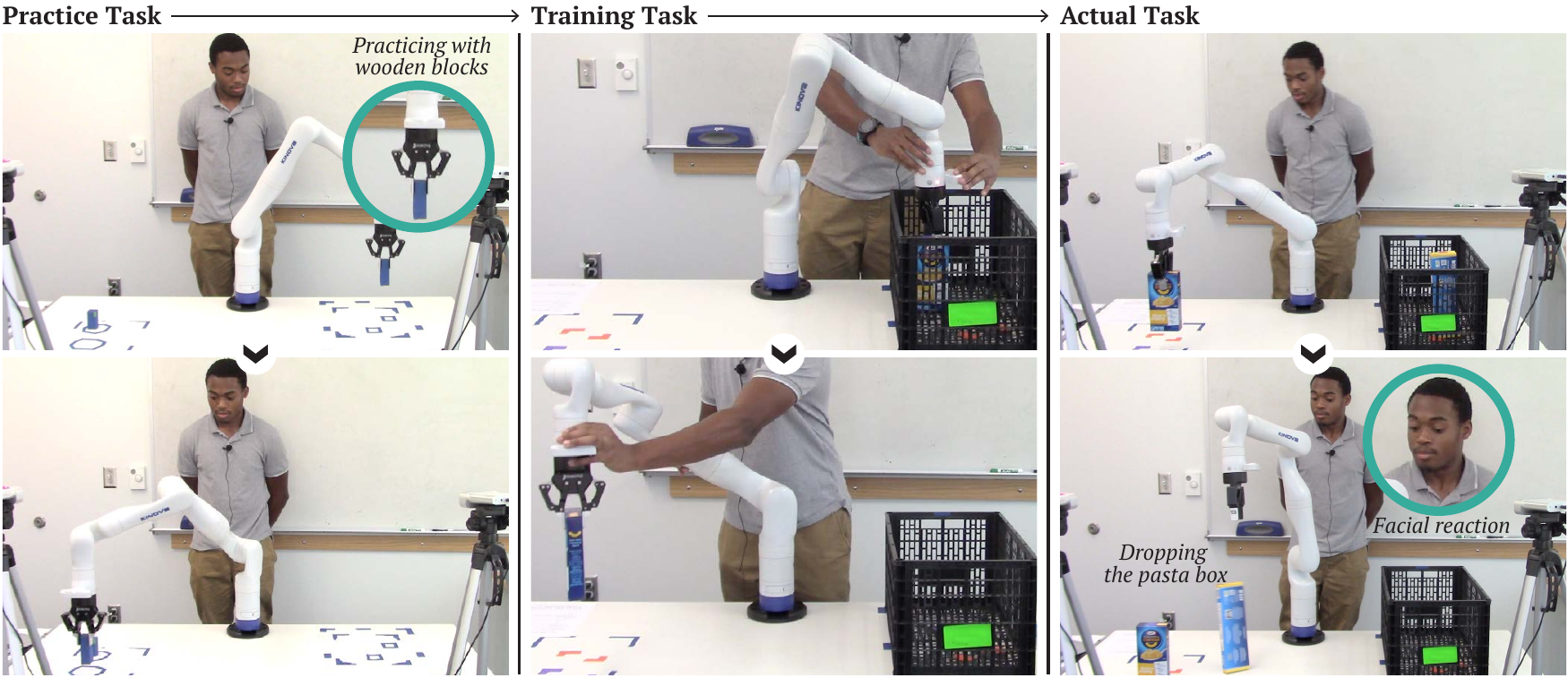}
\caption{Data collection study workflow. The practice task had the robot pick-and-place blocks from pre-designated positions, which was comparable to the actual task. Participants worked with the robot to train it in a pick-and-place task and then have it execute unpacking boxes from a crate. Participants kinethestically demonstrated to and programmed the robot by waypoint, using Box 1, the pick-and-place task needed to execute the actual task (Training Task). For the actual task, the robot  executed a grocery unpacking task. The robot first unpacked Box 1 and then simulated the generalization to unpack Box 2, during which it dropped the box (pre-programmed error).}
\label{fig:task-workflow}
\end{figure*}

In this work, we investigated whether it is possible to detect unexpected robot errors during physical interactions with a non-humanoid robot using facial action units (AUs). Our investigation consisted of (1) a data collection study to understand how people's AUs may manifest in response to technical robot errors in a non-social setting and (2) a data-driven model to allow for timely detection and localization of robot errors based on observed AUs.

\subsection{Data Collection}
Our data collection study was contextualized in a programming by demonstration (PbD) scenario in which participants provided task demonstrations through kinesthetic teaching using a Kinova Gen3 robot arm. This setup allowed participants to establish accurate mental models of the robot's capabilities and treated the participants as the task ``experts,'' similar to real-world applications. Throughout the study, a real-time data collection system logged facial AUs exhibited by participants in response to robot movements and errors. In addition, the experimenter was not in the room with the participant during our study except to introduce the task and the PbD interface as the presence of an experimenter has been shown to increase implicit social signal quantity expressed~\cite{giuliani2015systematic}.

\subsubsection{Study Task}~\label{ssec:proced1}
Before the actual data collection, participants completed a practice task, seeking to reduce any possible novelty effect associated with the robot moving, thereby leveling out participants' reactions to ``normal'' robot movements. 
The practice task, similar to the actual task, involved picking and placing wooden blocks with the robot executing the task without errors (see practice task in Fig.~\ref{fig:task-workflow}). 
Participants had the option to repeat the practice task until they were confident that they could program the robot and that the robot executed what they programmed.

For the actual data collection (see Fig.~\ref{fig:task-workflow} for training and actual task workflow), participants were asked to ``program'' the robot to unpack two pasta boxes from a crate and place each of them in predefined locations on a table.
The robot was ``trained'' with one of the boxes and executed the pick-and-place with both. 
The box the robot trained with was placed first without error before the robot ``generalized'' and performed a pick-and-place with the other box. 
During the execution of the second box's pick-and-place, we inserted a pre-programmed error---the robot dropping the box before it reached its goal---for the participant to observe and react to. 
Unbeknownst to the participants, their robot's training had no effect on the robot's behavior as its movements were pre-programmed by the experimenter.

\subsubsection{Study Procedure} 
After consenting to partake in the study, participants were informed about the task and taught how to program the robot. 
The participants then conducted the practice task. 
Once done, the experimenter confirmed participants' confidence in programming the robot and introduced the actual task. 
The actual task included the drop error embedded in robot execution. 
After the task, participants were asked to fill out a questionnaire about whether they witnessed an error, its severity, and basic demographics. 
The experimenter then debriefed the participants and informed them of the involved deception. The study lasted about $30$ minutes and was approved by our institutional review board (IRB). Participants were compensated \$8 for study completion.

\subsubsection{Study Systems}
In support of our data collection, we developed two systems to allow the experimenter to 
(1) operate the robot as in a Wizard of Oz (WoZ) paradigm~\cite{riek2012wizard} and track the study progress from a different room and 
(2) collect video and AU data. 
The first system was a PbD interface which served as a simulated programming environment, provided a sequence of instructions to the participants, and allowed the experimenter to oversee the task status for WoZ operations.
The second system logged user AUs while the robot carried out tasks.
Below, we provide detailed descriptions of the two systems.

\textit{A Simulated Programming System.} 
We created a 2D Unity application to increase the realism of our PbD scenario, hoping to lead participants to believe that they were actually programming the robot even though all of the robot's actions were actually pre-programmed. 
As a simulated programming system, the Unity application walked participants through a sequence of task steps. 
For the PbD portion of the task, the application provided the user a UI to add/delete waypoint and gripper commands as they kinethestically programmed the robot through the pick and place motions. 
On subsequent steps, the application informed participants about the task layout, robot movements, and then robot task execution. 
In addition, as the user progressed through the task steps, the application sent the experimenter relevant signals for WoZ operations (e.g., when to trigger robot movements and turn on the data collection system).

\begin{figure}[b]
  \includegraphics[width = \columnwidth]{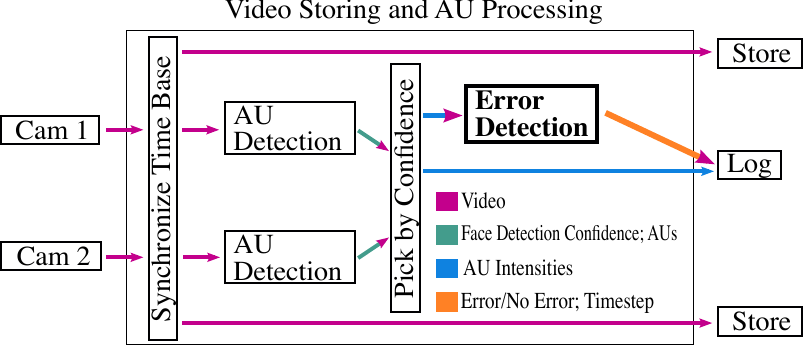}
\caption{Diagram of the real-time data collection platform. The bolded lines and box show how the error detection algorithm is integrated into the platform for the case study.}
\label{fig:data-collection-platform}
\end{figure}

\textit{A Data Collection System.} 
We constructed a data collection system (Fig.~\ref{fig:data-collection-platform}) that took in live video (30fps) and processed it in real-time to log AU occurrence and intensity for each timestep, where a timestep was 1/3 second.
Our system was built on top of Microsoft's Platform for Situation Intelligence (\textbackslash psi)~\cite{bohus2021platform}, which allowed us to synchronize multiple device inputs to a common time base, collect data, and add our real-time detection algorithm. 
We used OpenFace~\cite{baltruvsaitis2016openface} to extract 17 AUs. 
To allow the participants to have full freedom of body and head movement around the robot during the study, our system took input from two cameras strategically placed such that the system could maximize facial detection confidence regardless of participant position. The system then simultaneously stored the recordings and piped the videos to two instances of our AU detection component, one for each camera. These components computed AU occurrences and intensities, along with facial detection confidence. At each time step, the data from the component with higher facial detection confidence was logged. If neither components' facial detection confidence was above 50\%, then the AU calculation was considered inaccurate and zeros were logged for the AU metrics.

\begin{figure*} \vspace{2.5mm}
\centering
  \includegraphics[width = \textwidth]{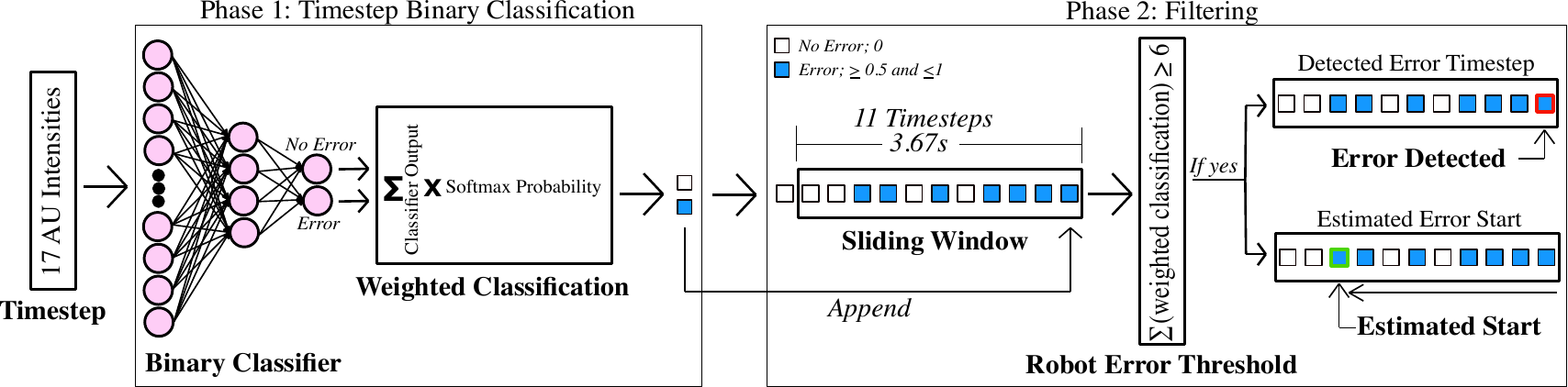}
\caption{Diagram of our detection algorithm. The input is the AU intensity of each timestep and the output, when the error is detected, is the timestep of error detection and the algorithm's estimated error start. Phase 1 consists of a 2-layer neural network that classifies each frame and the output of that is weighted by the softmax probability (classification confidence). The weighted classifications (no error: 0, error: $\geq 0.5$ and $\leq 1$) are the inputs for the sliding window of size 3.67s. If the window's convolution is greater than or equal to 6, then an error has been detected. The algorithm then traces back through the window to find the earliest timestep that is marked as an error and denoted as the estimated error start.}
\label{fig:algorithm}
\end{figure*}

\subsubsection{Participants}
Twenty-three participants were recruited for the study. Among all the participants, two participants exhibited strong robot novelty effects despite the practice task (i.e., consistently reacting to the robot picking up an object) and two exhibited no visible reaction to the robot errors. We considered these four participants special cases and did not include them in training our detection algorithm; however, we included an evaluation of our algorithm on these cases. 
Consequently, the resulting training data consisted of 11 females and 8 males with ages ranging from 18 to 39 ($M = 23.4, SD = 4.6$). Participants' experience with robots and technology was assessed through three questions on a 5-point scale (Cronbach's $\alpha = 0.83$); they had low-medium prior experience, $M = 2.93, SD = 0.89$.


\subsubsection{Dataset}


In total, we amassed 25.05 min of video of which 3.03 min consisted of participants' responses to the drop error through this data collection study. All 23 participants said that they did witness an error; however, two of them had no visible reaction. 
On average, the drop error's severity was evaluated as medium-high severe, $M = 4.91, SD = 1.70$, with 1 being low and 7 is high severity. 
In addition, we had each participant rate to what extent they thought the error was their fault as a metric to determine participant's confidence in programming the robot and perception of their mental model of the robot's capabilities. 
Across all participants, they minimally blamed themselves for the error, $M = 2.00, SD = 1.12$. 

\begin{figure*} \vspace{1mm}
\centering
  \includegraphics[width = \textwidth]{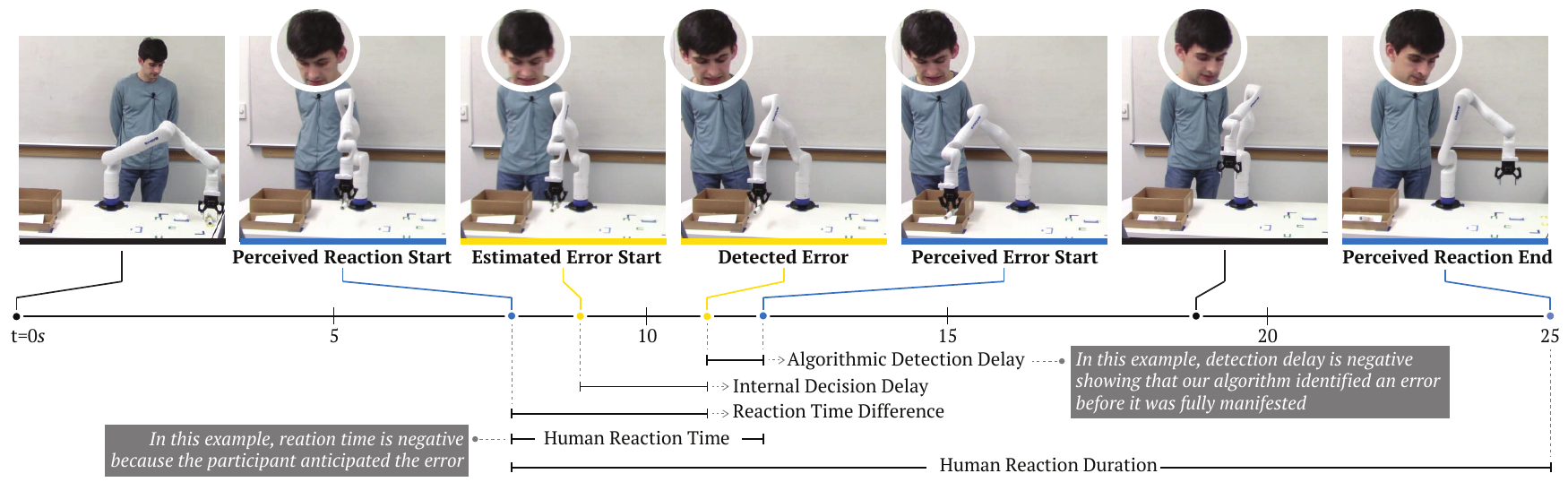}
\caption{Case study example of one participant's reaction to and our algorithm's error detection of a sorting error. The participant reacted to the error before the coder's perceived error start and so our algorithm was able to detect the error before then. The blue represents the coded timesteps (left to right): start reaction, perceived error start, and end reaction. The yellow indicates (left to right): our algorithm's estimated error start and robot error detected.}
\label{fig:error-detection}
\end{figure*}

In addition, two independent coders annotated the videos frame by frame to help quantify human response to robot errors in terms of reaction time and duration (See Fig.~\ref{fig:error-detection} for a visual representation):
\begin{itemize}
    \item \textbf{Human Reaction Time} (seconds). This metric quantifies how fast a participant reacted to the robot's error. It is defined as the difference between \textit{user reaction start} and \textit{perceived error start}. The user reaction start is defined as the time in which the first sign of any visible change in a participant's face happened. The perceived error start time is defined as when the coder was completely certain that the error was occurring, because some errors were slow to unfold. 
    The average reaction time of the participants to the error was 0.5s ($SD = 0.68$).
    
    \item \textbf{Human Reaction Duration} (seconds). This metric quantifies how long a participant reacted to the robot's error. It is defined as the difference between \textit{user reaction start} and \textit{user reaction end}---when the participant's face/behavior returned to their norm as seen through video. The average reaction duration was 11.78s ($SD = 7.08$).
\end{itemize}


We further examined if AU intensities during error and no-error instances were significantly different. 
Since we had heavily imbalanced data for error and no-error instances and potential unequal variance in the corresponding intensities, we used Welch's t-tests to compare the AU intensities.
Our analysis revealed that there was a statistically significant difference in intensities between error and no-error frames for 16 action units, except for AU\_4 (brow lowerer), illustrating the discriminative potential of AUs in characterizing the manifestation of an unexpected robot error.

\subsection{Modeling Action Units for Error Detection}

We designed an algorithm that capitalizes on the discriminative potential of AUs to detect and localize unexpected robot errors. Rather than building a ``complete'' error detector, our goal in this work is to explore the possibility of automatic recognition of robot errors using facial cues in non-social settings. We, however, explored various modeling methods including bidirectional LSTM, anomaly detection using an autoencoder, and SVM. Ultimately, due to the small size of our dataset and large imbalance in error versus non-error instances, we chose a simple, yet sufficiently robust method.
Our detection algorithm takes 17 AU intensities as inputs  at each timestep and outputs the timestep number at which an error is detected.  
The detection algorithm consists of two phases: 
(1) weighted binary classification and 
(2) sliding window filtering (Fig.~\ref{fig:algorithm}). 

\textit{Phase 1: Weighted Binary Classification.}
At each time step, 17 AU intensities are fed into a two-layer neural network (input: 17, hidden: 4, output: 2) that conducts timestep-by-timestep classification of participant expressions and outputs a binary classification of error versus no error.
In training this binary classifier, we accounted for the large imbalance in the number of error versus error-free timesteps by randomly undersampling the error-free timesteps every training epoch so that their counts matched those of the error timesteps. 
Moreover, each time step was treated as its own data point, and the temporal dimension was not preserved; this decision was made in part because of the small size of the training dataset.
Two independent coders annotated ground truth for each time step. 
The classification output from the network was then weighted by the classification confidence (softmax probabilities) and then summed to minimize the impact of out-of-distribution samples and misclassifications; prior work has shown this weighted approach to be effective~\cite{hendrycks2016baseline}. 

\textit{Phase 2: Sliding Window Filtering.}
To reduce spurious error classifications (e.g., any quick movements the participants made such as twitch), we employed a sliding window filter after the weighted binary classification. The sliding window was $3.67s$ long which translates to 11 timesteps. 
As every new weighted classification (ranging from 0 to 1 with 0 being no error) at each timestep was outputted, the window slid over by one timestep to include the most recent output. 
If the convolution of the window was at least $6$ (empirically tuned), then the algorithm outputted that a robot error had occurred. 

When an error was detected, we considered the \textit{estimated error start} as the earliest timestep in the window that was classified as an error and the \textit{detected error timestep} as the newest timestep of the latest error classification.
If the \textit{estimated error start} or \textit{detected error timestep} was within one timestep of a previously \textit{ detected error timestep}, then the present error was considered part of the former error.
We integrated this two-phase detection algorithm into our data collection system (Fig.~\ref{fig:data-collection-platform}) with AU intensities at each timestep inputted into the algorithm in real time. 
In production, our detection system could be run in real-time.

\subsection{Model Evaluation}
Typical ML metrics, such as accuracy and F1-score, are not representative of this algorithm's performance for two reasons: (1) errors are sparse in our dataset and 
(2) the classification accuracy of individual time step classification is not as important as sequential clustering of error time steps. 
Therefore, in evaluating our algorithm performance, we focused on the following metrics  (See Fig.~\ref{fig:error-detection} for a visual representation): 

\begin{itemize}
    \item \textbf{Algorithmic Detection Delay} (seconds). This metric measures the root mean squared error of the delay between \textit{algorithmic error detection} and \textit{perceived error start} (as annotated by independent coders).

    \item \textbf{Reaction Time Difference} (seconds). This metric represents the robot mean squared error of the time difference between \textit{algorithmic error detection} and coded \textit{user reaction start}.
    
    \item \textbf{Internal Decision Delay} (seconds). This metric computes the average difference between the algorithm's \textit{detected error timestep} and \textit{estimated error start}.

    \item \textbf{False Positive Rate}. This metric quantified the average number of false positives logged per trial. 
    A false positive is defined as when the algorithm's output timesteps (detected error timestep and estimated error start) do not overlap with the coded participant's reaction to the robot error.

\end{itemize}

We used leave-one-out cross validation with the collected data to evaluate the family of models used (feasibility of detecting errors), which allowed for a comprehensive evaluation of the model and also tested for overfitting due to the small size of the data sets. 
Our error detection and localization algorithm had an average \textit{algorithmic detection delay} of 3.25s and a \textit{reaction time difference} of 3.10s. 
The system's \textit{internal decision delay} was on average 2.37, $SD = 0.50$. 
In addition, the \textit{false positive rate} was 0.61 per trial, $SD = 0.78$, and false negative rate was 0 per trial. 
Table~\ref{table:results} summarizes results of our system evaluation.

\begin{table*} \vspace{2.5mm}
    \centering
    \caption{Summary of results for the two collected datasets. The grayed rows in the table show an additional analysis of the case study dataset as broken down by error type.}
    \includegraphics[width = \textwidth]{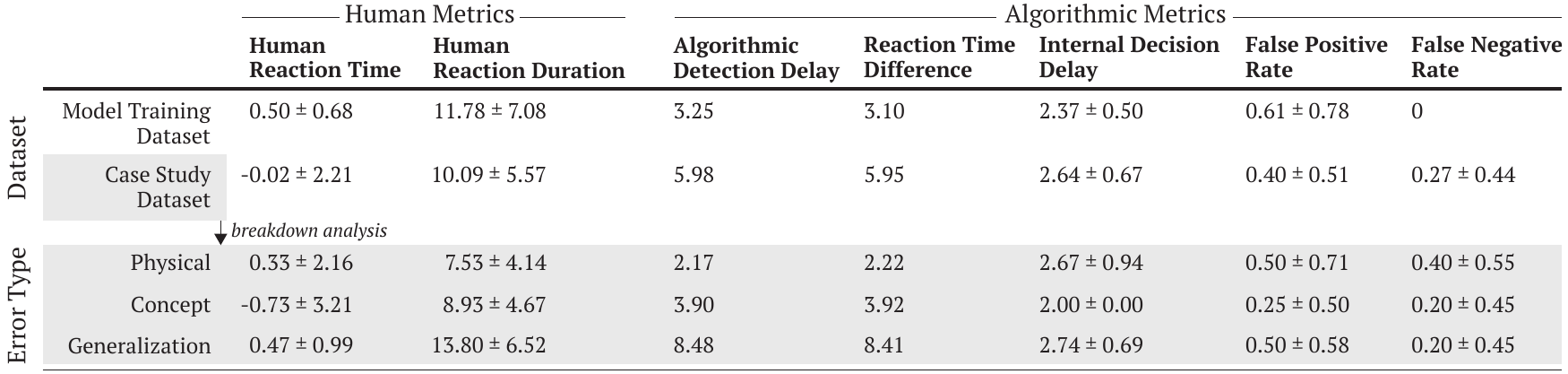}
        
    \label{table:results}
\end{table*}

\section{Real-time Error Detection and \\Model Generalization}
To explore how well our method for real-time error detection generalized to different tasks and error types, we ran a case study evaluating our algorithm against  different tasks and errors than the data collection study. 


\subsection{Task and Procedure}

\begin{table}
    \caption{Three error categories experienced by participants during the case studies of which each has two possible robot errors.}
    \includegraphics[width = \columnwidth]{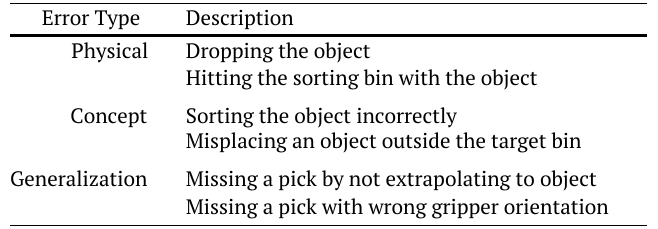}
    \label{table:errors}
\end{table}

We contextualized our case study in a sorting task, where participants demonstrated to the robot how to sort PVC pipes and joints into differently labeled bins. 
Similar to the data collection study, participants were given a practice task (sorting blocks by color) to familiarize themselves with the robot and the programming system. 
The task allowed participants to establish an understanding of the robot's capabilities to pick and place objects. 
In addition to demonstrating pick-and-place operations, we had the participants ``teach'' the robot the concept of sorting by having them show the robot's wrist cameras the object and placing it in the appropriate bin. 
The programming and teaching did not effect the robot's behavior; the robot errors were again pre-programmed.

For the main task, each participant experienced three different errors in three interaction rounds, where each round consisted of sorting three different objects into bins. 
In each round, the participant programmed the robot with only one of the objects, and the robot ``generalized'' to the other two objects when executed. 
In order to reset the participant's mental models between each round, we had them ``add'' onto the original robot program by programming the robot again with the object it just had made a mistake with. That object was then the first sorted by the robot in the following round to confirm successful program (i.e., reset mental model).

Participants were randomly assigned errors from three categories (\textit{physical}, \textit{concept}, and \textit{generalization} errors), one from each in a random order. 
Table~\ref{table:errors} lists the possible errors the participants could witness, the majority differing from the error seen in the data collection study. 
These error categories were chosen to represent the three aspects of the task: the robot was ``programmed'' for a pick-and-place (physical), was taught the notion of sorting into bins (concept), and extrapolated those two lessons towards new objects (generalization). 
All of the errors, except for the drop error, are considered to be \textit{predictable} errors meaning that the errors initially make slow perceptible changes in trajectory before they fully unfold. 

For this case study, the error detection algorithm, only trained on the data from  the data collection study,  was integrated into the data collection system (Fig.~\ref{fig:data-collection-platform}); the AUs were directly piped to the detection algorithm to allow for indication of when robot errors happened based on human reaction in real time.
The overall procedure for this case study was comparable to that in the data collection study. This study lasted around one hour and participants were compensated \$16.

\subsection{Participants}
Five participants (three female and two male) were recruited for this case study. Their ages ranged from 22 to 29 ($M = 25.2, SD = 3.11$) and had medium prior experience with technology and robots, $M = 3.00, SD = 0.85$.

\subsection{Case Analysis}
The case study dataset consisted of 23.47 min of robot interaction video, where 2.61 min involved facial reactions to errors. 
The average error severity (1 is low and 7 being high severity) for all errors rated by the participants was medium severe ($M = 4.33, SD = 1.60$) where physical ($M = 4.00, SD = 2.00$), concept ($M = 3.80, SD = 1.48$) and generalization ($M = 4.60, SD = 0.55$) errors were all considered medium severe. 
All participants confirmed that they saw the robot make an error and attributed low blame (1 being low and 7 is high blame) to themselves, $M = 3.00, SD = 1.85$. 


On average, the \textit{human reaction time} was 0.02s before coded \textit{perceived error start}, $SD = 2.21$, indicating that the participants reacted before the error was fully manifested.
The average \textit{human reaction duration} was 10.09s, $SD = 5.57$. When we look at participants' reactions to the different error types (Table \ref{table:results}), we see that, on average, participants reacted to concept errors before the \textit{perceived error start} and the \textit{human reaction duration} for the generalization errors were longer than that of for physical and concept errors.
Furthermore, we ran the best resulting model, only trained on the data collection dataset, with the case study data as a test set (generalizability of use cases). 
The algorithm detected, in real-time, errors with an \textit{algorithmic detection delay} of 5.98s and a \textit{reaction time difference} of 5.95s. It had an \textit{internal decision delay} of 2.64s, $SD = 0.67$. 
In addition, the \textit{false positive rate} was 0.40, $SD = 0.51$, and a false negative rate of 0.27, $SD = 0.44$.
If we were to separate the model's evaluation by error type, the results reveal that the algorithm was able to timely detect the physical and concept errors but was delayed about twice as long as for detecting generalization errors. See Table~\ref{table:results} for a summary of the statistics for this dataset.

\section{Discussion}

\begin{figure} \vspace{2.5mm}
  \includegraphics[width = \columnwidth]{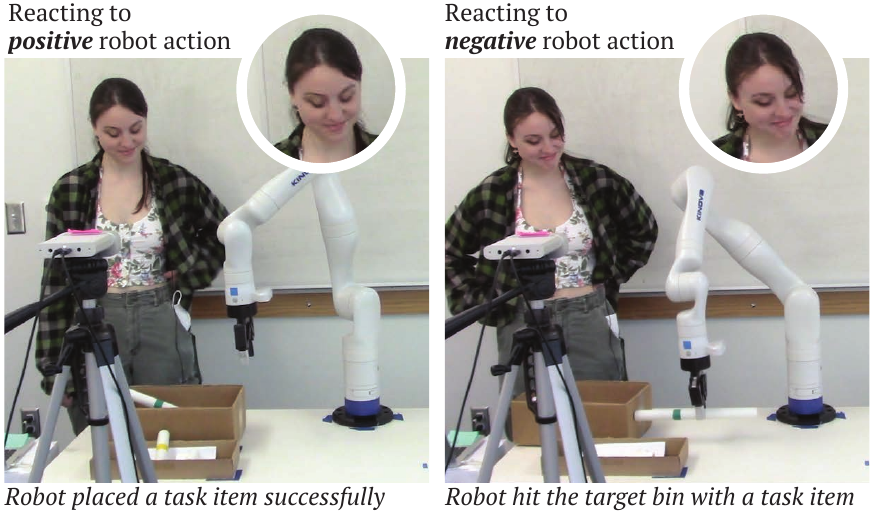}
\caption{Example of similar reaction to positive and negative robot actions. The left image (positive robot action) shows the robot moving away from a bin after it properly and correctly placed a pipe in it. The right image (negative robot action) shows the robot making an error by pushing the same bin with the same pipe.}
\label{fig:posNegAction}
\end{figure}

In this paper, we demonstrate leveraging natural human response to unexpected robot errors to automatically detect them in real-time. Our results show that it is possible to detect and temporally localize errors within reasonable accuracy and timeliness using AUs and that this algorithm can be generalized to a different task and to different types of errors. More importantly, this approach is not person-specific.

\subsection{Error Detection Through Social Responses}
Robot errors likely and immediately elicit social responses. Of all 28 participants in both studies, only two did not have visible reactions to errors. 
In addition, participant reaction times were, on average, within a second of the perceived error start for both studies. 
By relying on implicit, social reactions to recognize robot errors, our approach does not lose noticeable time to ``wait'' for the user's explicit responses, as evidenced by \textit{algorithmic detection delay} and \textit{reaction time difference}, which are  within 0.15s of each other.
On the flip side, our approach in principle is only as fast as detectable social responses.

However, if the error is predictable by the human---namely if the error causes gradual noticeable deviations in behavior/trajectory before it fully unfolds---it is possible for the person to observe and react before the \textit{perceived error start}. 
In our case study, some errors were predictable, and when examining participants' reaction times to predictable vs. non-predictable errors, we found that, on average, they reacted before the \textit{perceived error start} for predictable errors, $M = -0.03, SD = 2.38$, and after for non-predictable errors $M = 0.33, SD = 0$. Thus, the algorithm has the ability to take advantage of the early reaction and indicate that the robot is making a mistake before the \textit{perceived error start}. Fig.~\ref{fig:error-detection} illustrates an example of such a detection.

\subsection{Reliability, Customizability, and Generalizability}
All of our results were consistent despite the fact that no two participants reacted the same, even when the errors were the same. Localization and detection of the robot error for our data collection study was 3.25s delayed; for the case study, the \textit{algorithmic detection delay} was 2.73s longer as compared to the data collection study. Nevertheless, our approach was able to reliably detect different types of errors in situations in which it was not trained.

We explored this discrepancy further by training the algorithm on one of the trials for each person in the case study, tailoring it to each person, before testing it on the remaining trials for that person. This fine-tuning improved \textit{algorithmic detection delay} and \textit{reaction time difference} by 0.74s and 0.60s, respectively. These results are similar to what we see in human-human interaction where people have a harder time (are less accurate) decoding meaning from facial expressions with strangers than with friends~\cite{sternglanz2004reading}. Consequently, this finding shows the potential for improving performance through fine-tuning detection, which would be useful for longer-term interactions with the same person. 

Our algorithm detects large changes in a person's facial reactions from their norm and then makes the assumption that those changes are due to errors. This is, however, not always the case. Indeed, we observed an average \textit{false positive rate} of about 0.51 per trial over both studies. All of the false positives detected were a consequence of the participants reacting to different robot actions, such as placing an object correctly or relief from correct operation after an error. The algorithm cannot tell the difference between a reaction due to a positive or negative robot action; \emph{it needs context} (Fig.~\ref{fig:posNegAction}). In human-human interaction, facial expressions are inherently ambiguous without context~\cite{hassin2013inherently} and the same facial expressions in different cultures could mean different things~\cite{crivelli2016reading}. 
However, it is important to note that some visible reactions are not detected as errors by the algorithm.

\subsection{Limitations}\label{ssec:limit}
One limitation of our approach is that detection can only occur if the user reacts and within a reasonable time frame. We tested our algorithm on the two trials from the data collection study where the participants had no reaction, and the algorithm failed to detect the errors. 
On the opposite side, if the user reacted to everything (mostly due to the novelty effect), then the algorithm would constantly generate false positives, as shown by testing our algorithm on two trials from the data collection study where the participants exhibited strong novelty effects (false positive rate: $M = 2, SD = 0$). 
In addition, the algorithm could not discern differences between reactions to positive robot actions and negative robot actions (errors). 
Another limitation is related to reliable facial detection. For example, if the individual's face is obscured (e.g., by their hand) then our approach would not be able to detect AUs. Moreover, we found that if the person were to remove the obstruction, then there would be a jump in AU intensity, triggering a false positive. 
Finally, we had limited datasets; therefore, the algorithm was not necessarily trained with a fully comprehensive array of human responses to robot errors. In addition, we were not able to implement this approach using other modeling methods that require larger training sets.

\subsection{Implications for Human-Robot Collaboration}
Our findings illustrate the feasibility of detecting various types of errors in different tasks in real-time using natural responses during human-robot interaction. This work is the first step towards robust error detection, key to successful error management. While our approach can detect errors across different users, it also has the potential to be tailored to an individual for more accurate error detection and localization.

Future work will focus on fully integrating the algorithm into an autonomous robotic system that will perform error recovery after detection. In addition, we look to collect more data, improve the detection algorithm by exploring different social signals, and add context to improve the false positive rate. We should consider how this social signal-based approach may be used with other error detection methods to improve detection reliability and flexibility. We also want to explore automatic classification of error severity through social signals as that could provide information for appropriate recovery strategies and help us understand the impact of those errors.

\section*{ACKNOWLEDGMENT}
We thank Kaitlynn Pineda for performing the video coding for the errors and reactions. This work was supported by the Johns Hopkins University Institute for Assured Autonomy and the National Science Foundation award \#2143704. 

\balance
\bibliographystyle{IEEEtran}
\bibliography{references}

\end{document}